\documentclass[conference]{IEEEtran}
\IEEEoverridecommandlockouts
\usepackage{cite}
\usepackage{amsmath,amssymb,amsfonts}
\usepackage{algorithmic}
\usepackage{graphicx}
\usepackage{textcomp}
\usepackage{xcolor}
\usepackage{float}
\usepackage{url}
\renewcommand{\baselinestretch}{0.99}

\newcommand\Wg[1]{\colorbox{lightgray}{$#1$}}

\makeatletter
\let\MYcaption\@makecaption
\makeatother

\usepackage[font=footnotesize]{subcaption}

\makeatletter
\let\@makecaption\MYcaption
\makeatother

\def\BibTeX{{\rm B\kern-.05em{\sc i\kern-.025em b}\kern-.08em
    T\kern-.1667em\lower.7ex\hbox{E}\kern-.125emX}}
\begin{document}

\title{Balancing Sparse RNNs with Hyperparameterization Benefiting Meta-Learning \\
 \thanks{Research was performed using computational resources supported by the Academic and Research Computing group at Worcester Polytechnic Institute.}
 }


\author{\IEEEauthorblockN{Quincy Hershey}
\IEEEauthorblockA{\textit{Data Science} \\
\textit{Worcester Polytechnic Institute}\\
Worcester, United States \\
qbhershey@wpi.edu}
\and
\IEEEauthorblockN{Randy Paffenroth}
\IEEEauthorblockA{\textit{Mathematical Sciences, Computer Science, and Data Science} \\
\textit{Worcester Polytechnic Institute}\\
Worcester, United States \\
rcpaffenroth@wpi.edu}
}

\maketitle

\begin{abstract}
This paper develops alternative hyperparameters for specifying sparse Recurrent Neural Networks (RNNs). These hyperparameters allow for varying sparsity within the trainable weight matrices of the model while improving overall performance. This architecture enables the definition of a novel metric, hidden proportion, which seeks to balance the distribution of unknowns within the model and provides significant explanatory power of model performance. Together, the use of the varied sparsity RNN architecture combined with the hidden proportion metric generates significant performance gains while improving performance expectations on an a priori basis. This combined approach provides a path forward towards generalized meta-learning applications and model optimization based on intrinsic characteristics of the data set, including input and output dimensions. 
\end{abstract}


\section{Introduction}\label{sec:introduction}

Selection and specification of neural networks remains an actively studied arena \cite{YANG2020295, 9718485, yu2020hyperparameteroptimizationreviewalgorithms}. Practitioners traditionally follow a pathway of first selecting a model architecture suited to the general characteristics of the data set. Afterwards, model hyperparameters are often optimized through a process involving repeated training runs and cross-validation \cite{refaeilzadeh2009cross}. The search for optimal hyperparameters is often costly as the total potential combinations grow exponentially with the number of hyperparameters and model effectiveness can often only be judged after training. This paper addresses these challenges by developing novel hyperparameters and a method of balancing the distribution of trainable unknowns on an a priori basis. The resulting generalized architecture improves model performance across diverse tasks \cite{hershey2023exploring, 10459970}.

Recurrent neural networks (RNNs) \cite{rumelhart1986learning, Sherstinsky_2020} were first developed as an approach to sequential machine learning tasks. Recent research has brought a new focus to RNNs configured with sparse trainable weight matrices. Sparsely configured networks have demonstrated stable performance versus traditional dense networks while also benefiting from parameter efficiency in terms of scale \cite{DBLP:journals/corr/abs-1711-02782, Hoefler_sparse, dodge2019rnn}. Existing research on sparse RNNs has focused on achieving sparsity through pruning methods \cite{frankle2019lottery, https://doi.org/10.48550/arxiv.1909.03011, https://doi.org/10.13140/rg.2.2.30539.20004} or utilizing adaptive connectivity \cite{Liu2021}. 

\begin{figure}[htbp]
  \centering
  \includegraphics[width=0.5\textwidth]{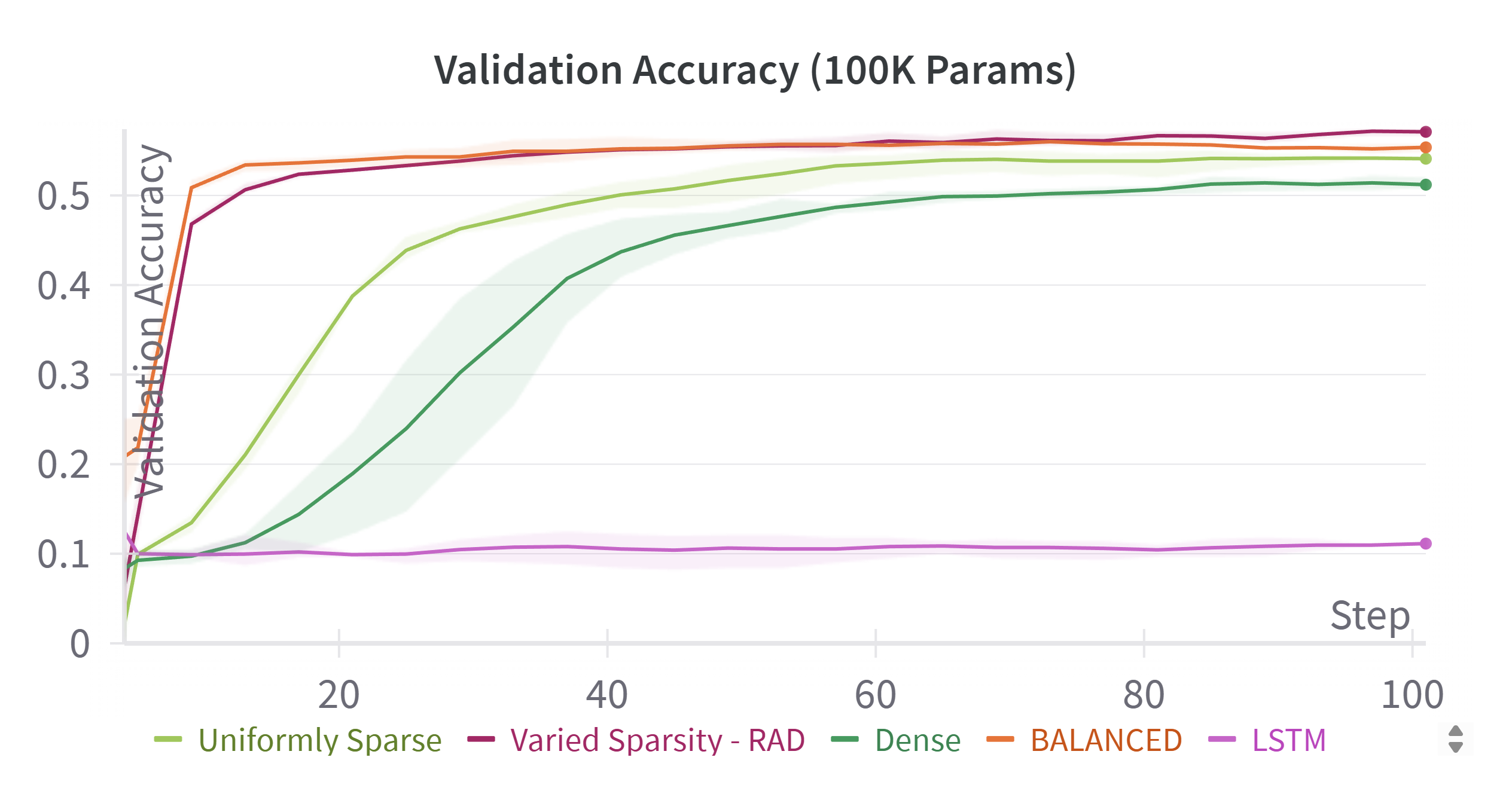}
  \caption{The RNNs in red were specified using hyperparameter tuning while the RNNs in orange were specified a priori by balancing the number of unknowns. Both the RNNs in red and orange allow the sparsity of the unknowns to vary within regions of their weight matrices, outperforming the 100\% dense RNNs (dark green), uniformly 20\% sparse RNNs (light green) and LSTMs (purple). Models have $\sim$100k trainable weights on Random Anomaly Detection (RAD) for 25 epochs.}
  \label{fig:RA_Balanced}
\end{figure}

More recently, research has shown clear benefits to stability and performance by instantiating sparse RNNs using uniform sparsity throughout the weight matrices \cite{hershey2023exploring, 10459970}. This paper illustrates the limitations of that method while also finding significant performance gains by taking a novel approach allowing sparsity to vary within regions of the weight matrices.

Contrary to common perceptions, sparsely configured RNNs have been demonstrated to be a generalized class of networks \cite{hershey2024rethinkingrelationshiprecurrentnonrecurrent}, capable of representing a wide range of other architectures, including MLPs, on sequential and non-sequential data. It is important to highlight that the interpretation of sparse RNNs as a generalized network architecture is a mathematically provable construct \cite{hershey2024rethinkingrelationshiprecurrentnonrecurrent} with novel implications, including that MLPs may be viewed mathematically as a strict subset of sparse RNNs. By developing this concept using novel hyperparameters which are more predictive of model performance, inroads are developed towards meta-learning across diverse problem sets. Specifically, the first step of model selection based on the data set may instead be deferred to a generalized class of networks with a universal set of shared hyperparameters. The second step of costly model tuning may be abbreviated or eliminated, with algorithms better positioned to predict optimal specifications based on task characteristics.

This paper explores and develops alternative hyperparameters for sparse RNNs. These hyperparameters vary sparsity within the trainable weight matrices to influence the allocation of weights based on the input and output dimensions of a task. Model performance is studied using thousands of training runs across a series of reinforcement learning (RL) and anomaly detection tasks. \textbf{The result introduces a novel metric, \textit{hidden proportion}. The hidden proportion metric reflects the influence of sparsity within weight matrices, seeking to balance the arrangement of unknowns within the model while improving explanatory power (Fig. \ref{fig:Introchart}), performance predictions (Fig. \ref{fig:RA_RandomForestMother}) and overall performance (Fig. \ref{fig:RA_Balanced}).}

\hfill

\noindent \textbf{This paper makes the following novel contributions:}

\begin{itemize}
\item Varied sparsity RNNs using effective hyperparameters improve model specification and performance.
\item Input and output dimensions of the task combine with the \textit{hidden proportion} metric to significantly improve understanding of performance.
\item The \textit{hidden proportion} metric improves a priori estimation of optimal hyperparameters.
\item This paper creates a pathway for a generalized approach to meta-learning across diverse tasks.
\end{itemize}

\section{Sparse Block RNNs}\label{sec:method_iterative_representation}

This paper follows the notation outlined in \cite{hershey2024rethinkingrelationshiprecurrentnonrecurrent} and extended in \cite{reinforcementRNNs}. We reproduce some of the derivations from \cite{reinforcementRNNs} to make the current text stand alone. The basic notational conventions are provided for reference in Table \ref{tab:notation}.

\begin{table}[ht]
  \centering
  \begin{tabular}{|c|c|}
    \hline
    \textbf{Notation} & \textbf{Description} \\
    \hline
    $x$ & Scalar \\
    \hline
    $\mathbf{x}$ & Column vector \\
    \hline
    $|\mathbf{x}|$ & Number of dimensions in $\mathbf{x}$ \\
    \hline
    $X$ & Matrix \\
    \hline
  \end{tabular}
  \caption{Notation for scalars, vectors, and matrices.}
  \label{tab:notation}
\end{table}

\noindent When defining a typical RNN, related vector values exist for each step $t$ in a given data sequence. Notably, the input $\mathbf{x}_{t} \in \mathbb{R}^{|\mathbf{x}|}$ is used to generate the hidden state $\mathbf{h}_{t}\in \mathbb{R}^{|\mathbf{h}|}$ which serves as the input for the model prediction $\mathbf{y}_{t}\in \mathbb{R}^{|\mathbf{y}|}$. In practice, a series of $n$ input sequences may be processed in batches forming a matrix $X_{t} \in \mathbb{R}^{|\mathbf{x}| \times n}$ resulting in a traditional RNN defined as

\begin{align}
  H_{t+1} &= \sigma(W_{h,x} \cdot X_{t+1} + W_{h,h} \cdot H_{t} + \mathbf{b}) \nonumber \\
  Y_{t+1} &= W_{y,h} \cdot H_{t+1} 
  \label{eqn:Rnneqbatch}
\end{align}

where the trainable weight tensors of an RNN are defined as, $W_{y,h} \in \mathbb{R}^{|\mathbf{y}| \times |\mathbf{h}|}$, $W_{h,x} \in \mathbb{R}^{|\mathbf{h}| \times |\mathbf{x}|}$, $W_{h,h} \in \mathbb{R}^{|\mathbf{h}| \times |\mathbf{h}|}$, and $\mathbf{b}^{|\mathbf{h}|}$.

To begin constructing a block function equivalent, a \emph{block non-linear function} is first defined in Eqn. \eqref{eqn:BNLO}, where $f_{i,j}$ is free to be any function and the composition operator $\circ$ now combines standard functional composition with the dot-product as 

\begin{equation}
\begin{bmatrix}
f_{0,0} & f_{0,1} \\
f_{1,0} & f_{1,1} \\
\end{bmatrix}
\circ
\begin{bmatrix}
A \\
B
\end{bmatrix}
=
\begin{bmatrix}
f_{0,0}(A) + f_{0,1}(B) \\
f_{1,0}(A) + f_{1,1}(B) \\
\end{bmatrix}.
\label{eqn:BNLO}
\end{equation}

In the specific case where the underlying functions are linear, Eqn. \eqref{eqn:BNLO} simply results in a standard block matrix dot-product. Focusing on the affine case, we define $f_{i,j}(A) = W_{i,j} \cdot A + b_{i,j}$. The block non-linear functions are then combined as

\begin{equation}
  F = \begin{bmatrix} f_{h,x} & f_{h,h} & f_{h,y} \\ f_{y,x} & f_{y,h} & f_{y,y} \\ \end{bmatrix}. 
  \label{eqn:iterativeY}
\end{equation}

\noindent By first concatenating, $X_{t+1}$, $H_{t}$, and $Y_{t}$ into a single input matrix $Z_{t+1}$ then defining $\Sigma$ as 

\begin{equation}
  Z_{t+1} = \begin{bmatrix} X_{t+1} \\ H_{t} \\ Y_{t} \end{bmatrix} \qquad and \qquad
  \Sigma = \begin{bmatrix} \sigma & 0 \\ 0 & I \end{bmatrix},
  \label{eqn:Z_Sigma}
\end{equation}   

\noindent the batch RNN in Eqn. \eqref{eqn:Rnneqbatch} can be written as a generalized block RNN from Eqn. (7) in \cite{reinforcementRNNs} as shown below

\begin{equation}
\begin{split}
\begin{bmatrix}
H_{t+1} \\
Y_{t+1}
\end{bmatrix} 
 & =
\Sigma \circ F \circ Z_{t+1} \\
 & = 
\begin{bmatrix}
 \sigma(f_{h,x}(X_{t+1}) + f_{h,h}(H_{t}) + f_{h,y}(Y_{t})) \\
 f_{y,x}(X_{t+1}) + f_{y,h}(H_{t}) + f_{y,y}(Y_{t}) \\
 \end{bmatrix}.
\end{split}
\label{eqn:genRNN}
\end{equation}

Finally, we define the \textbf{\textit{weight space}} as the set of all trainable weight tensors $W_{i,j}$ from the affine functions, which underlie the block function $F$ in Eqn. \eqref{eqn:iterativeY}. RNNs with varied sparsity are constructed by specifying each $W_{i,j}$ individually as a sparse matrix with differing degrees of sparsity. It follows by example that a block sparse RNN containing an arrangement of 20\% sparsity blocks and fully dense blocks (100\% sparsity) can be specified as 

\begin{equation}
\Sigma \circ \begin{bmatrix} \Wg{f_{100}} & f_{20} & \Wg{f_{100}} \\ f_{20} & \Wg{f_{100}} & f_{20} \\ \end{bmatrix} \circ \begin{bmatrix} X_{t+1} \\ H_{t} \\ Y_{t} \end{bmatrix}
\label{eqn:sparseexample}
\end{equation}

\noindent where $f_{20}$ indicates the weight matrix within the function contains approximately 20\% randomly distributed trainable weights \footnote{The bias term is always dense and trainable.} with the remainder initialized as $0$ entries having no gradient. The $\Wg{f_{100}}$ entries have dense underlying weight matrices containing 100\% randomly initialized trainable weights.  \textbf{Notably, each} $f_{20}$ \textbf{or} $\Wg{f_{100}}$ \textbf{represents a distinct instance of each function with its own unique parameter matrix.}

\section{Data sets}\label{sec:method_datasets}

Experiments on sequential data sets were conducted on an anomaly detection task and several offline reinforcement learning tasks using PyTorch \cite{NEURIPS2019_9015} Lightning \cite{Falcon_PyTorch_Lightning_2019} with results tracking in Weights and Biases \cite{wandb}. Python implementations of the code that produced the results in this paper can be found at \url{https://github.com/qbhershey/iterativenn}. Results were contrasted between these tasks leading to novel insights into sparse RNN performance. Ultimately, these insights demonstrate the substantial explanatory power of the hidden proportion metric in determining model behavior. The data sets underlying these experiments are outlined in Table \ref{tab:datasets} and in further detail below.

\begin{table}[htbp]
\caption{Sequential Training Tasks}
\begin{center}
\begin{tabular}{|c|c|c|c|c|c|}
\hline
\textbf{} & \textbf{Total} & \textbf{Series} & \textbf{Input} & \textbf{Output} & \textbf{} \\
\textbf{Task} & \textbf{Series} & \textbf{Length} & \textbf{Dimension} & \textbf{Dimension} & \textbf{Reward} \\
\hline
AH$^{\mathrm{a}}$ & 5000 & 200 & 46 & 26 & Dense \\
\hline
RAD$^{\mathrm{c}}$ & 30000 & 9 & 2500 & 10 & n/a \\
\hline
\multicolumn{6}{l}{$^{\mathrm{a}}$ Adroit Hammer Expert (AH)} \\
\multicolumn{6}{l}{$^{\mathrm{c}}$ Random Anomaly Detection (RAD)} \\
\end{tabular}
\label{tab:datasets}
\end{center}
\end{table}

\subsection{Offline Reinforcement Learning}\label{sec:method_sequential_rl}

Experiments involving offline reinforcement learning data sets and environments were conducted using the Gymnasium \cite{Towers_Gymnasium} and Gymnasium Robotics \cite{Gymnasium-Robotics_Contributors_Gymnasium-Robotics_A_a_2022} repositories. These data sets and environments are publicly available and maintained, allowing for benchmarking and reproducible outcomes. Reinforcement Learning (RL) \cite{kaelbling1996reinforcement} tasks feature an agent network taking actions within an environment based on the preceding observed state or states with the goal of maximizing the reward. Within \textit{offline} reinforcement learning tasks, the network is provided the observations with corresponding actions and rewards from an expert agent from which it attempts to learn. The loss function in offline reinforcement learning seeks to minimize deviation versus the expert agent's behavior. Offline RL experiments used the Minari \cite{Minari_Contributors_Minari_A_dataset_2023} data sets maintained by the Farama Foundation. The original environments remain accessible through the repository, so final testing was conducted using rewards achieved by the trained agent in the actual environment. The high-level characteristics of two primary data sets are shown in Table \ref{tab:datasets}.

The Adroit Hammer Expert (AH) data set is based on the Adroit Hammer \cite{rajeswaran2017learning} task which involves a simulated robotic hand grabbing a hammer and driving randomly placed nails into a horizontal board within a given episode. Of the 5,000 episodes, 4,500 were used for training while 500 were used for validation. Points are scored based on the degree of task completion while a time penalty is applied to encourage speed and efficiency. The Adroit Hammer task is the easiest of the supervised RL tasks given more training episodes, dense reward structure and overall simplicity.


\subsection{Sequential Random Anomaly Detection}\label{sec:method_sequential}

Experiments were also conducted using the Random Anomaly Detection (RAD) data set \cite{hershey2023exploring}. The inclusion of the anomaly detection data set in addition to reinforcement learning allows examination of network behavior across two highly dissimilar tasks. Each task presents very different requirements for data retention or ``memory'' within the networks, as well as very different input and output dimensions and differing ultimate objectives. 

The Random Anomaly Detection task was created from the MNIST \cite{lecun2010mnist} (Modified National Institute of Standards and Technology) database. MNIST is a common choice for machine learning exercises, with the content of each image being a hand written single digit integer and ground truth label. In the Random Anomaly Detection task \cite{hershey2023exploring, 10459970}, each sequence contains one image unlike the others based on grouped image transformations.  The task requires the network to predict the indexed location of the anomaly after processing each sequence. Each series is created by randomly selecting a collection of $n=9$ different hand-written images of the same numerical digit. For $n-1$ of those images, the same randomly selected combination of image transformations is applied, while the remaining image at a random location in the sequence will be the anomaly \textit{and} have a different randomly selected combination of transformations applied. A key feature of this data set is the disparity between the input dimension of 2,500 and the output dimension of 10 (when n=9). The input and output dimensions influence the geometry of the weight matrices, impacting performance and becoming a key focus of discussion in the sequel.

\section{Balancing network architectures}\label{sec:balanced_networks}

The performance of the sparse block RNNs was tested on the RL tasks \cite{reinforcementRNNs} using mixed configurations of 20\% sparse and dense blocks within the weight space. The best model performance was achieved with

\begin{equation}
\Sigma \circ \begin{bmatrix} \Wg{f_{100}} & f_{20} & f_{20} \\ \Wg{f_{100}} & \Wg{f_{100}} & f_{20} \\ \end{bmatrix} \circ \begin{bmatrix} X_{t+1} \\ H_{t} \\ Y_{t} \end{bmatrix}.
\label{eqn:VariedSparsity}
\end{equation}

\begin{figure}[htbp]
  \centering
  \includegraphics[width=0.5\textwidth]{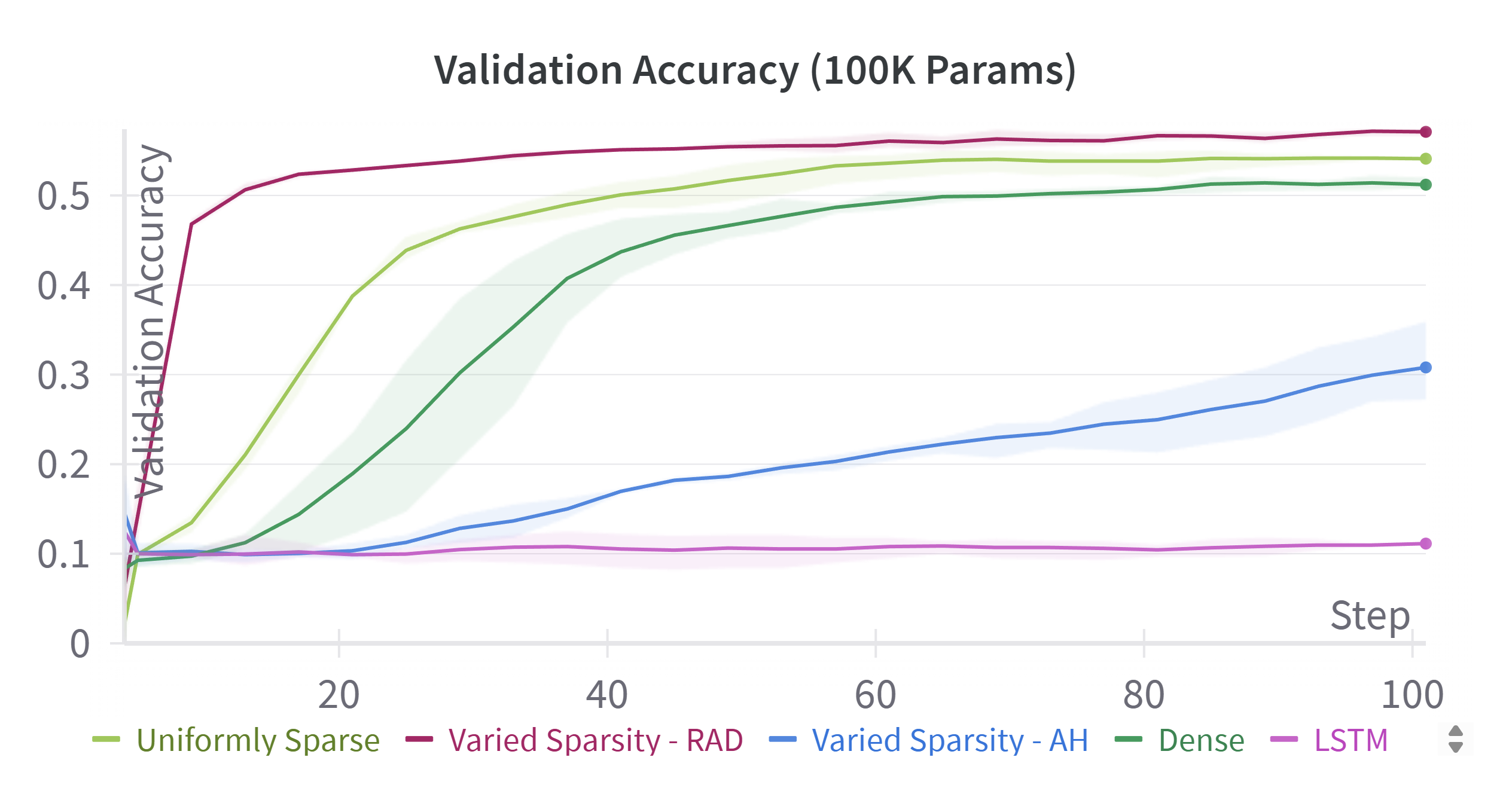}
  \caption{Varied sparsity models trained on Random Anomaly Detection (RAD) and benchmarked using validation accuracy versus training steps. Results are shown for varied sparsity RNNs initially configured for the Adroit Hammer task (blue) from Eqn. \eqref{eqn:VariedSparsity} but trained on RAD and for networks reconfigured for RAD (maroon) from Eqn. \eqref{eqn:RA_Varied}. By comparison, results are shown for 100\% dense RNNs (dark green), uniformly 20\% sparse RNNs (light green) and LSTMs (purple).}
  \label{fig:RA_Varied_Losses}
\end{figure}

Experiments were then conducted using the Random Anomaly Detection (RAD) task with models limited to approximately 100,000 trainable weights. The objective of this exercise was to examine whether the optimal sparsity arrangement from Eqn. \eqref{eqn:VariedSparsity} would prove to be universal or task specific. In contrast to the Adroit Hammer task, deploying this configuration on the RAD task resulted in stark model underperformance in Fig. \ref{fig:RA_Varied_Losses}. In the case of RAD, the uniformly 100\% dense and uniformly 20\% sparse networks both performed well. However, the specific varied sparsity configuration from the Adroit Hammer task in Eqn. \eqref{eqn:VariedSparsity} which mixed sparse blocks with dense blocks dramatically underperformed networks composed of either entirely sparse or dense weights. Surprisingly, reversing the optimal configuration from Eqn. \eqref{eqn:VariedSparsity} as  

\begin{equation}
\Sigma \circ \begin{bmatrix} f_{20} & \Wg{f_{100}} & \Wg{f_{100}} \\ f_{20} & f_{20} & \Wg{f_{100}} \\ \end{bmatrix} \circ \begin{bmatrix} X_{t+1} \\ H_{t} \\ Y_{t} \end{bmatrix}
\label{eqn:RA_Varied}
\end{equation}

resulted in a varied sparsity RNN that outperformed all other configurations on RAD. Adding to this, the varied sparsity network arrangement that performed best on RAD was the worst performing on the Adroit Hammer task and vice versa. As a reminder, these two configurations are inverse arrangements of one another.

The characteristics driving this task specific behavior appear to be driven by the disproportionate relationship between the 2,500 dimensions of the input and the 10 dimensions of the output on RAD. This disparity skews the geometry of the weight space, influencing the proportional allocation of parameters between those which act on the input versus the hidden state. In other words, one of the primary effects of sparsity is to alter the dimensions or "\textit{aspect ratio}" of the overall weight space dimensions. Importantly, the dimensions of weight matrices relating to the input and output are fixed, only allowing the size of a block to expand \textit{linearly} along the hidden state dimension. The lone exception is the $|\mathbf{h}| \times |\mathbf{h}|$ central block of the weight space which expands \textit{exponentially}, receiving the prior hidden state as input and outputting data to the next hidden state. 

Consider the example of a network specified for RAD with approximately 100,000 trainable parameters. As discussed in Sec. \ref{sec:method_iterative_representation} and shown again in Eqn. \eqref{eqn:Z_dimensions}, the input matrix $Z_{t+1}$ contains the input matrix $X_{t+1}$ of fixed dimension $|\mathbf{x}| \times n$ concatenated with the hidden state matrix $H_t$ of dimension $|\mathbf{h}| \times n$ where $|\mathbf{h}|$ is a specified characteristic of the model, concatenated with the output matrix $Y_t$ of fixed dimension $|\mathbf{y}| \times n$. Of these dimensions, only $|\mathbf{h}|$ is permitted to vary freely, aside from $n$, which does not influence the dimensions of the weight space. In scenarios where fully dense networks are specified at a given number of parameters, then $|\mathbf{h}|$ also becomes fixed as well.

\textbf{However, when the network functions become sparse arrangements, the dimension of} $|\mathbf{h}|$ \textbf{now varies based on network specification. It is clear by looking at both} Eqn. \eqref{eqn:Z_dimensions} \textbf{and} Eqn. \eqref{eqn:trainable_calc} \textbf{that the $|\mathbf{h}| \times |\mathbf{h}|$ central block of the top row of the weight space expands in two dimensions, while the other functions are fixed in one or both dimensions and thus expand linearly or not at all. Moreover, through the use of sparsity as a hyperparameter, the proportion of parameters allocated between the} $H_t$ \textbf{and} $X_{t+1}$ \textbf{inputs can be allocated in a parameter efficient manner, leading to improved network stability.}

\begin{equation}
\label{eqn:Z_dimensions}
Z_{t+1} = \begin{bmatrix} X_{t+1} \in \mathbb{R}_{|\mathbf{x}| \times n} \\ H_t\in \mathbb{R}_{|\mathbf{h}| \times n} \\ Y_t\in \mathbb{R}_{|\mathbf{y}| \times n} \end{bmatrix}
\end{equation}

\noindent Trainable parameters ($P_T)$ per affine block function can be calculated where $S(f_{sparsity})$ is the sparsity of a function as 

\begin{equation} \label{eqn:P_T}
  P_T(f_{sparsity}) = |output| \times |input| \times S(f_{sparsity})
\end{equation}

In a fully dense configuration, the network sparsity functions are shown on the left in Eqn. \eqref{eqn:trainable_calc} with dimensions determined by the maximum network size, shown in the center in Eqn. \eqref{eqn:trainable_calc}. The trainable parameters are calculated per block on the right of Eqn. \eqref{eqn:trainable_calc} using the pointwise multiplication Hadamard Product $\odot$ with Eqn. \eqref{eqn:P_T} as $P_T(RNN) =$

\begin{equation} 
\begin{split}
\begin{bmatrix} | \mathbf{h}| \times | \mathbf{x}| & | \mathbf{h}| \times | \mathbf{h}| & | \mathbf{h}| \times | \mathbf{y}| \\ | \mathbf{y}| \times | \mathbf{x}| & | \mathbf{y}| \times | \mathbf{h}| & | \mathbf{y}| \times | \mathbf{y}| 
\end{bmatrix} 
\odot \\
\begin{bmatrix} S(f_{100}) & S(f_{100}) & S(f_{100}) \\ S(f_{100}) & S(f_{100}) & S(f_{100}) \\ \end{bmatrix}.
\end{split}
\label{eqn:trainable_calc}
\end{equation}

\begin{figure}[H]
   \centering
   \begin{minipage}[b]{0.3\textwidth}
     \centering
     \includegraphics[width=\textwidth]{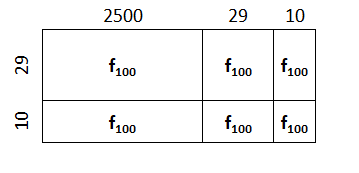}
     \subcaption{}
     \label{fig:trainable_dense_example1}
   \end{minipage}
   \hfill
   \begin{minipage}[b]{0.3\textwidth}
     \centering
     \includegraphics[width=\textwidth]{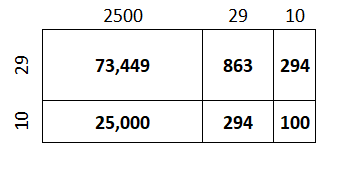}
     \subcaption{}
     \label{fig:trainable_dense_example2}
   \end{minipage}
    \caption{The number of unknowns per block for a dense RNN with $\sim 100k$ unknowns on the Random Anomaly Detection task. The number of unknowns per block can be calculated from Fig. \ref{fig:trainable_dense_example1} using Eqn. \eqref{eqn:trainable_calc} with the resulting arrangement of unknowns illustrated in Fig. \ref{fig:trainable_dense_example2}. }
    \label{fig:trainable_dense_example}
\end{figure}

\noindent Using the specific case of a fully dense RNN on the Random Anomaly task, with maximum trainable parameters of 100,000, the arrangement of unknowns is illustrated in Fig. \ref{fig:trainable_dense_example} as

\begin{figure}[H]
   \centering
   \begin{minipage}[b]{0.3\textwidth}
     \centering
     \includegraphics[width=\textwidth]{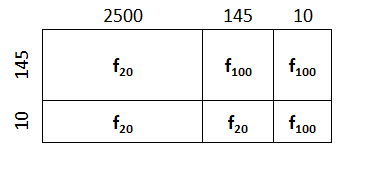}
     \subcaption{}
     \label{fig:trainable_ra_example1}
   \end{minipage}
   \hfill
   \begin{minipage}[b]{0.3\textwidth}
     \centering
     \includegraphics[width=\textwidth]{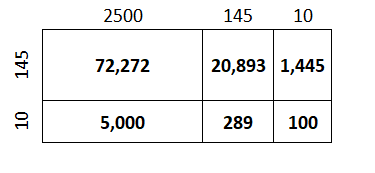}
     \subcaption{}
     \label{fig:trainable_ra_example2}
   \end{minipage}
    \caption{The number of unknowns per block for the best performing varied sparsity RNN arranged as in from Eqn. \eqref{eqn:RA_Varied} with $\sim 100k$ unknowns on the Random Anomaly Detection task. The number of unknowns per block can be calculated from Fig. \ref{fig:trainable_ra_example1} using Eqn. \eqref{eqn:trainable_calc} with the resulting arrangement of unknowns illustrated in Fig. \ref{fig:trainable_ra_example2}. }
    \label{fig:trainable_ra_example}
\end{figure}

\noindent Now, continuing with a 100,000 parameter network on the Random Anomaly task, we explore the parameter allocation of the varied sparsity network from Eqn. \eqref{eqn:RA_Varied} which had worked best on the problem with the arrangement of unknowns illustrated in Fig. \ref{fig:trainable_ra_example}.

\begin{figure}[H]
   \centering
   \begin{minipage}[b]{0.3\textwidth}
     \centering
     \includegraphics[width=\textwidth]{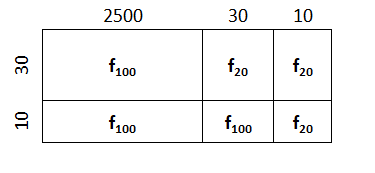}
     \subcaption{}
     \label{fig:trainable_rl_example1}
   \end{minipage}
   \hfill
   \begin{minipage}[b]{0.3\textwidth}
     \centering
     \includegraphics[width=\textwidth]{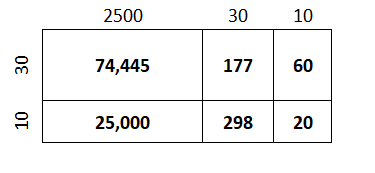}
     \subcaption{}
     \label{fig:trainable_rl_example2}
   \end{minipage}
    \caption{The number of unknowns per block for the worst performing varied sparsity RNN arranged as in from Eqn. \eqref{eqn:VariedSparsity} with $\sim 100k$ unknowns on the Random Anomaly Detection task. The number of unknowns per block can be calculated from Fig. \ref{fig:trainable_rl_example1} using Eqn. \eqref{eqn:trainable_calc} with the resulting arrangement of unknowns illustrated in Fig. \ref{fig:trainable_rl_example2}. }
    \label{fig:trainable_rl_example}
\end{figure}

\noindent Here it is clear that the proportion of weights that receive $X_{t+1}$ as input has decreased, allowing more weights to be allocated to receiving the prior hidden state as input. The resulting network becomes more balanced. Lastly, we show the result for the worst performing network configuration Eqn. \eqref{eqn:VariedSparsity} with the arrangement of unknowns illustrated in Fig. \ref{fig:trainable_rl_example}.

\textbf{When considering the weight space the most important indicator of network performance was the proportion or percentage of parameters in the block row calculating $H_{t+1}$, which receive $H_t$ as input.} We call this the \textit{\textbf{hidden proportion}}. Taking the output from Fig. \ref{fig:trainable_ra_example2}, the proportion of parameters informing $H_{t+1}$, which also receive $H_t$ as input can be calculated as the \textbf{\textit{hidden proportion}} with

\begin{equation}
20,893 / (72,272+20,893+1,445) = 0.22 \; or \; \mathbf{22\%} .
\end{equation}

\begin{figure}[htbp]
  \centering
  \includegraphics[width=0.5\textwidth]{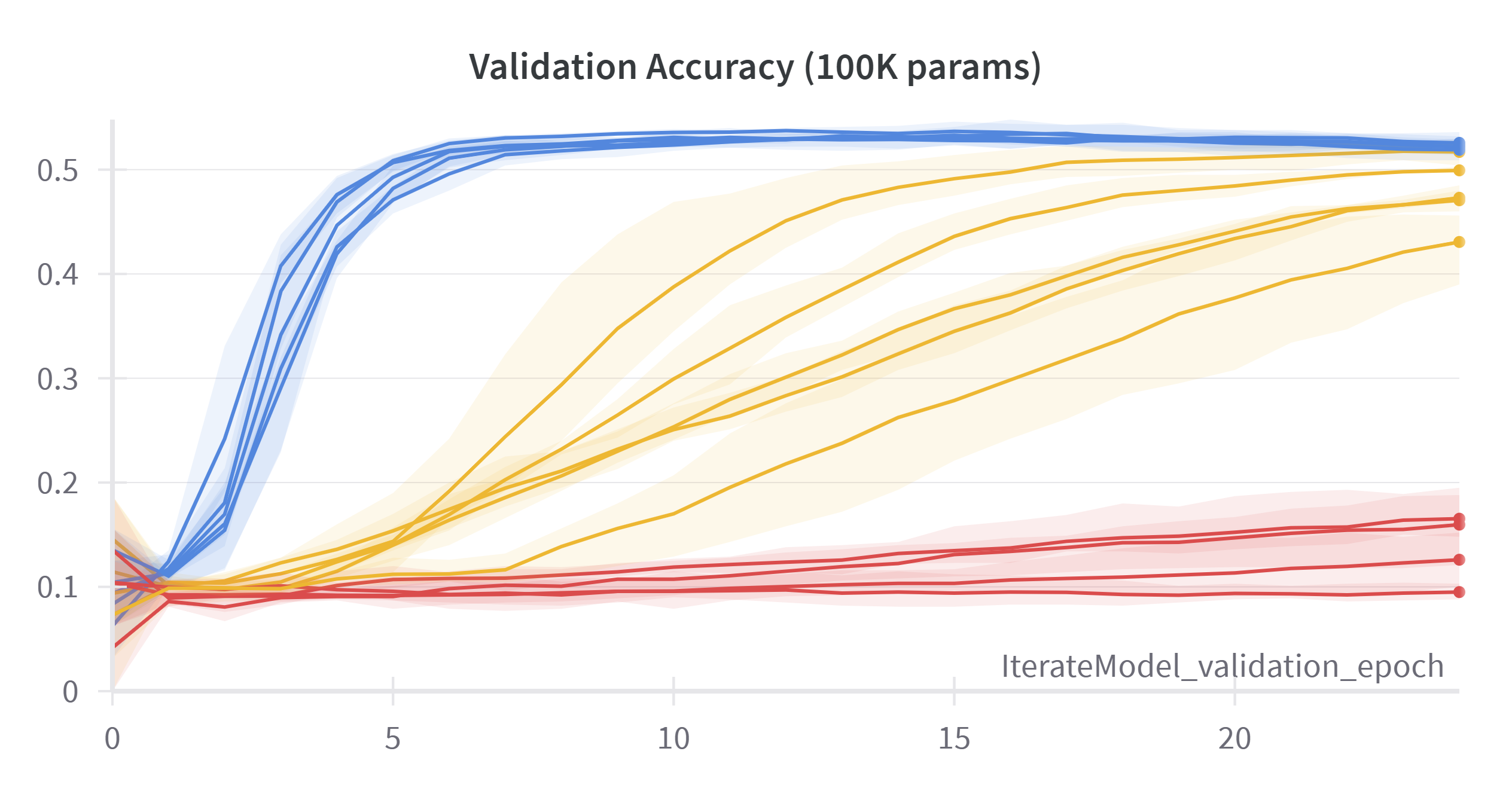}
  \caption{\textbf{Color coding reflects the value range of \textit{Hidden Proportion}}, a single explanatory hyperparameter influenced by model sparsity. Each line reflects a sparse RNN configuration training on the Random Anomaly task. Validation accuracy (y-axis) is presented by epoch (x-axis).}
  \label{fig:Introchart}
\end{figure}

The explanatory value of the hidden proportion metric is shown in Fig. \ref{fig:Introchart} which examines networks with $\sim 100k$ trainable parameters. The chart is color coded with the tight grouping of the best networks in blue having hidden proportion metric of 20-24\%. The middle grouping of dispersed performance networks colored yellow had a hidden proportion metric of 2-8\% and the worst performing networks in red had a hidden proportion metric of 0\% (reflecting rounding). In other words, trainable weights were allocated in a very imbalanced manner as the 2,500 input features versus 10 output features distorted the geometry of the weight space. The result in this extreme case was effectively no allocation of parameters towards memory retention. This weakness was shared by a comparably specified Long-Short Term Memory (LSTM) network \cite{6795963, Sherstinsky_2020}. The rigidity of the LSTM architecture in a weight constrained environment used nearly all the weights for the input and gating mechanisms, leaving effectively 0\% to memory retention. Importantly, the more evenly balanced a varied sparsity RNN was along the hidden proportion metric, the better it performed on the task in each case. 

Models were specified by varied sparsity hyperparameters using a simple calculation which evenly balanced the number of unknowns within the model using the hidden proportion metric. \textbf{This approach to model design requires only the input and output dimensions of the data set and uses no validation testing}. Using this method to ``balance'' the unknowns within a varied sparsity RNN with input of 2,500 features and output of 10 features resulted in a 47\%  hidden proportion using an 8\% sparse block as

\begin{equation}
\Sigma \circ \begin{bmatrix} f_{08} & \Wg{f_{100}} & \Wg{f_{100}} \\ \Wg{f_{100}} & \Wg{f_{100}} & \Wg{f_{100}} \\ \end{bmatrix} \circ \begin{bmatrix} X_{t+1} \\ H_{t} \\ Y_{t} \end{bmatrix}.
\label{eqn:RA_Balanced}
\end{equation}

The results for 100,000 parameter networks on RAD are shown in Fig. \ref{fig:RA_Balanced}, the network with balanced unknowns is in orange. Interestingly, this network outperforms all dense, uniformly sparse and LSTM configurations of similar size by a significant margin. The network with balanced unknowns also converged more rapidly than the varied sparsity network which had been tuned through extensive cross validation and testing. Validation accuracy declined slightly in later epochs as the model overfit and training continued, which would have been prevented with early stopping. \textbf{Ultimately, networks that have the distribution of unknowns balanced according to the hidden proportion metric outperformed all other configurations on both tasks (RL results to be published in \cite{reinforcementRNNs}) while bypassing the need for costly hyperparameter tuning.} As such, varied sparsity RNNs with concise hyperparameters demonstrate improved performance on an a priori basis. The topic of predicting model performance based on task characteristics is further explored in the next section.

\section{Parent and child networks}\label{sec:mother_child_networks}

There are several clear benefits that arise from viewing MLPs and RNNs as arising within a single generalized architecture as detailed in \cite{hershey2024rethinkingrelationshiprecurrentnonrecurrent}. Doing so allows for techniques, hyparameters and methodologies to be shared across a much wider array of problem sets, tasks and data types. Additionally, establishing a single predictive set of hyperparameters enables better a priori specification of optimal model architecture.

\subsection{Adroit hammer}\label{sec:mother_child_AH}

Considering the potential for improved hyperparameterizxation of generalized networks, we explore whether machine learning algorithms can accurately predict model performance on a given task given just the hyperparameters of that network. A data set was constructed using 2,083 separate training runs on the Adroit Hammer task. Each run consisted of 25 training epochs using randomly varied hyperparameters and with number of trainable weights left to vary. The network hyperparameters serve as the independent variable while the validation loss serves as the dependent variable, forming a data set relating hyperparameters to network performance. From there, simple machine learning algorithms were used to explore the data set characteristics.

\begin{figure}[htbp]
  \begin{subfigure}[b]{0.24\textwidth}
   \centering
   \includegraphics[width=\textwidth]{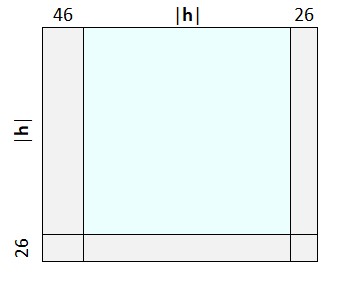}
   \caption{}
   \label{fig:AH_Illustrate_Blue}
  \end{subfigure}
  \hfill
  \begin{subfigure}[b]{0.23\textwidth}
   \centering
   \includegraphics[width=\textwidth]{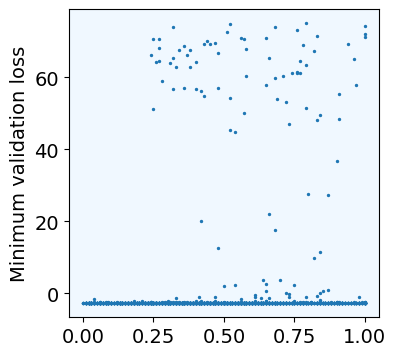}
   \caption{}
   \label{fig:AH_SparsityGrid}
  \end{subfigure}
  \caption{An illustration in Fig. \ref{fig:AH_Illustrate_Blue} of proportionally drawn weight space for the Adroit Hammer task partitioned into six block functions based on input and output. The blue portion of the block matrix in Fig. \ref{fig:AH_Illustrate_Blue} is isolated in Fig. \ref{fig:AH_SparsityGrid}, plotting the effect of sparsity in that block versus minimum validation loss for the Adroit Hammer task. As the block becomes more sparse (with sparsity approaching 0.0), the incidence of models failing to converge decreases. Each data point represents one trained network with varied unknowns.}
  \label{fig:AH_SparsityGrid_illustrate}
\end{figure}

The data set consists of 20 input dimensions comprised of the core hyperparameters which define a varied sparsity block RNN and one additional feature. The weight space was notationally divided among six regions based upon the input and output for the parameter blocks as shown in Fig. \ref{fig:AH_Illustrate_Blue}. The first nineteen of the input dimensions included the six means, six variances and six sparsities for each of these regions in addition to the model learning rate. The final feature was the hidden proportion metric. The data set target was comprised of the 25 validation losses captured at each epoch, with the minimum used for performance plots.




The hidden proportion and average model sparsity are indicative of model performance as shown in Table \ref{tab:AH_Hidden_Sparsity}. The hidden proportion metric represents the proportion of unknowns receiving the prior hidden state as input before informing the next hidden state. The minimum validation loss in the data set (summarized by mean) declines before reaching a bottom as the hidden  proportion approaches 0.5 and the networks become ``balanced''. The same is true as the overall networks become more sparse and fewer networks fail to converge. These two characteristics are interrelated, the relationship follows from the effect wherein average model sparsity is heavily influenced by the sparsity of the largest block. Increasing sparsity in the largest block tends to shift the allocation of weights into surrounding blocks, ultimately balancing the number of unknowns, which is reflected in the hidden proportion metric. As Fig. \ref{fig:AH_Illustrate_Blue} illustrates, the central block is disproportionately large owing to the small fixed input and output dimensions. As the central block and overall model become more sparse, the hidden proportion becomes more balanced in Table \ref{tab:AH_Hidden_Sparsity} and performance improves with reduced negative outliers.

\begin{table}[ht]
  \centering
  \begin{tabular}{|c|c|c|c|c|c|}
    \hline
    \textbf{Metric} & $\mathbf{<.2}$ & $\mathbf{.2-.4}$ & $\mathbf{.4-.6}$ & $\mathbf{.6-.8}$ & $\mathbf{>=.8}$  \\
    \hline
    Hidden Proportion & $0.071$ & $0.069$ & $\mathbf{0.068}$ & $uc$ & $uc$ \\
    \hline
    Model Sparsity & $\mathbf{0.068}$ & $uc$ & $uc$ & $uc$ & $uc$ \\
    \hline
  \end{tabular}
  \caption{Minimum validation loss (summarized by mean) for 2,083 models with varied unknowns on the Adroit Hammer task. \\ UC: unstable convergence, loss not meaningful.}
  \label{tab:AH_Hidden_Sparsity}
\end{table}

The central block often dominates the geometry of the weight space owing to the fixed proportion of the input and output dimensions. This relationship is shown again in Fig. \ref{fig:AH_SparsityGrid_illustrate} where Fig. \ref{fig:AH_Illustrate_Blue} illustrates a proportionally drawn weight space for the Adroit Hammer task partitioned into six block functions based on input and output. The blue portion of the block matrix in Fig. \ref{fig:AH_Illustrate_Blue} is isolated in Fig. \ref{fig:AH_SparsityGrid}, plotting the effect of sparsity in that block versus minimum validation loss for the Adroit Hammer task. As the block becomes more sparse (with sparsity approaching 0.0), the incidence of models failing to converge also decreases. Within the shaded block, sparsity below 30\% is associated with much more stable training and lower minimum validation losses with high explanatory power.

\subsection{Random anomaly}\label{sec:mother_child_RA}

A similar analysis was then conducted on the Random Anomaly task. A data set was once again formed using randomly generated hyperparameters to train models over 25 epochs. For this data set, 3,254 networks were trained with each restricted to approximately 50,000 total parameters. The resulting data set contained the same nineteen hyperparameters including the six means, variances and sparsities used for initial model parameterization and the model learning rate. The inclusion of the hidden proportion metric resulted in 20 total independent features and a target vector of 25 validation losses captured at each epoch through training.



\begin{figure}[htbp]
  \centering
  \includegraphics[width=0.5\textwidth]{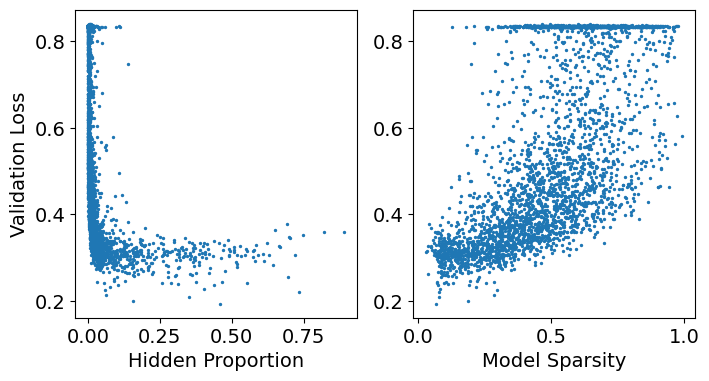}
  \caption{Performance of models with $\sim 50k$ unknowns on the Random Anomaly task with the \textbf{hidden proportion metric} (left) and average model sparsity (right) versus minimum validation loss. As the proportion of unknowns (left) which receive the prior hidden state as input \textbf{increases} away from 0.0, the frequency of networks with high validation loss (which fail to converge) decreases. Similarly, on the right, as the overall model becomes more sparse (with sparsity approaching 0.0), results improve with fewer networks failing to converge. Visualization of data from Table \ref{tab:RA_Hidden_Sparsity}.}
  \label{fig:RA_Hidden_Sparsity}
\end{figure}

As with the Adroit Hammer data set, the related characteristics of hidden proportion and model sparsity prove to be primary determinants of model performance, shown in Table \ref{tab:RA_Hidden_Sparsity} and Fig. \ref{fig:RA_Hidden_Sparsity}. Examining these characteristics reveals that in the Random Anomaly problem more sparsity also improves performance. However, the arrangement of the sparsity within the weight space proves vital and carries an inverse impact on the hidden proportion metric. As the model becomes more sparse (with sparsity approaching 0.0), the proportion of weights in the central block \textbf{increases} in Fig. \ref{fig:RA_Hidden_Sparsity}. This relates to the large input dimension of the data set which skews the geometry of the weight space. The block which receives $X_{t+1}$ as input and outputs to the hidden state is the largest in the model owing to the large input dimension of the data set. As a result, the sparsity of this block heavily influences overall model sparsity. As this disproportionately large block becomes more sparse, hidden proportion becomes more balanced as unknowns are allocated to receive $H_t$ as input and overall model performance improves. 

\begin{table}[ht]
  \centering
  \begin{tabular}{|c|c|c|c|c|c|}
    \hline
    \textbf{Metric} & $\mathbf{<.2}$ & $\mathbf{.2-.4}$ & $\mathbf{.4-.6}$ & $\mathbf{.6-.8}$ & $\mathbf{>=.8}$  \\
    \hline
    Hidden Proportion & $1.791$ & $\mathbf{1.351}$ & $1.352$ & $1.373$ & $1.431$ \\
    \hline
    Model Sparsity & $\mathbf{1.381}$ & $1.511$ & $1.810$ & $2.026$ & $2.091$ \\
    \hline
  \end{tabular}
  \caption{Minimum validation loss (summarized by mean) for 3,254 models with $\sim 50k$ unknowns on the Random Anomaly task. \\ Data visualized in Fig. \ref{fig:RA_Hidden_Sparsity}}
  \label{tab:RA_Hidden_Sparsity}
\end{table}

\begin{figure}[htbp]
  \begin{subfigure}[b]{0.24\textwidth}
   \centering
   \includegraphics[width=\textwidth]{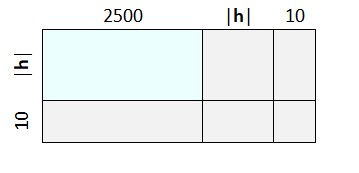}
   \caption{}
   \label{fig:RA_Illustrate_Blue}
  \end{subfigure}
  \hfill
  \begin{subfigure}[b]{0.23\textwidth}
   \centering
   \includegraphics[width=\textwidth]{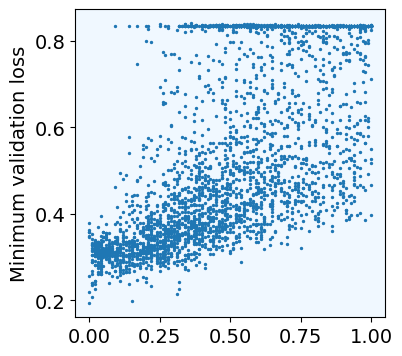}
   \caption{}
   \label{fig:RA_SparsityGrid}
  \end{subfigure}
  \caption{An illustration in Fig. \ref{fig:RA_Illustrate_Blue} of proportionally drawn weight space for the Random Anomaly task partitioned into six block functions based on input and output. The blue portion of the block matrix in Fig. \ref{fig:RA_Illustrate_Blue} is isolated in Fig. \ref{fig:RA_SparsityGrid}, plotting the effect of sparsity in that block versus minimum validation loss for the Random Anomaly task. As the block becomes more sparse (with sparsity approaching 0.0), the incidence of models failing to converge decreases. Each data point represents one trained network with $\sim$50k trainable weights.}
  \label{fig:RA_SparsityGrid_illustrate}
\end{figure}

As demonstrated in Fig. \ref{fig:RA_SparsityGrid_illustrate} the geometry of the weight space in Fig. \ref{fig:RA_Illustrate_Blue} is skewed such that the largest block is to the left of the central weight space block. The result is that as model sparsity increases, the central block grows fastest and a higher proportion of the total parameters are shifted to the central block, causing performance to rise as the network becomes more balanced. Just as before, a more balanced parameter allocation leads to stronger network performance and improved stability. Because the left block that is shaded blue is the largest, this influence of sparsity can most clearly be seen in that block's plot in the right portion of Fig. \ref{fig:RA_SparsityGrid}. As the block becomes more sparse (with sparsity approaching 0.0), the incidence of models failing to converge decreases.

\begin{figure}[htbp]
  \centering
  \includegraphics[width=0.5\textwidth]{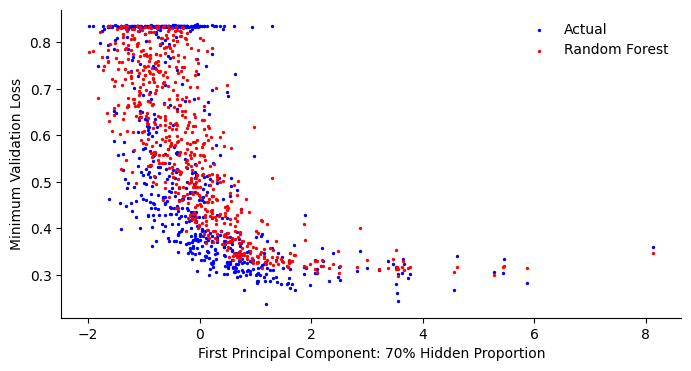}
  \caption{\textbf{A random forest predicts minimum validation loss using the \textit{Hidden Proportion} metric and model hyperparameters on an a priori basis}. This single metric reflects model structure and offers significant predictive power of model performance. Each data point represents a trained network.}
  \label{fig:RA_RandomForestMother}
\end{figure}

Combining these generalized networks with improved explanatory power from improved hyperparameterization lays the groundwork to explore networks capable of specifying optimal architectures on an a priori basis. In this scenario, a parent network may examine high-level data and then inform child model hyperparameter selection using predicted loss curves. Such a parent and child architecture would allow for improved meta-learning across tasks. As a first step towards exploring this concept on the Random Anomaly task, a neural network and random forest classifier were each trained on the normalized hyperparameter data set and the minimum validation loss through training of each network. The result is shown in Fig. \ref{fig:RA_RandomForestMother}. Here, a random forest classifier featuring 1,000 estimators and a max depth of 10 is able to accurately predict the performance of each network based on its hyperparameters against the validation set. The effectiveness of the random forest classifier illustrates the predictive power of these metrics in estimating model performance.

\section{Conclusion}\label{sec:conclusion}

Sparse RNNs create a generalized framework for analyzing neural network performance across diverse problems. This paper unveils a new set of hyperparameters suited for defining sparse RNNs with varying levels of sparsity within the weight space. The resulting networks demonstrate clear gains in parameter efficiency, stability and ultimately performance across a variety of tasks. Adopting this novel approach improves model specification and improves understanding of model performance through the use of the \textbf{\textit{hidden proportion}} metric. By improving understanding of model performance on sparse RNN architectures, the ability to specify optimal hyperparameters on an a priori basis lends itself to meta-learning applications. 

Further research efforts are focused on extending these findings directly into meta-learning applications across diverse problem sets. The logical next steps in this research involve developing a generalized approach to model specification tested on unseen tasks.


\bibliographystyle{unsrt}
\bibliography{references}

\end{document}